\PassOptionsToPackage{prologue,table,xcdraw}{xcolor}
\documentclass[sigconf]{acmart}
\usepackage{multirow}
\usepackage{subfigure}
\usepackage{diagbox}
\usepackage[table,xcdraw]{xcolor}
\usepackage{bbding}
\usepackage{pifont}

\copyrightyear{2023}
\acmYear{2023}
\setcopyright{acmlicensed}
\acmConference[MM '23] {Proceedings of the 31st ACM International Conference on Multimedia}{October 29--November 3, 2023}{Ottawa, ON, Canada.}
\acmBooktitle{Proceedings of the 31st ACM International Conference on Multimedia (MM '23), October 29--November 3, 2023, Ottawa, ON, Canada}
\acmPrice{15.00}
\acmISBN{979-8-4007-0108-5/23/10}
\acmDOI{10.1145/3581783.3612436}

\begin{document}

\title{Fearless Luminance Adaptation: A Macro-Micro-Hierarchical Transformer for Exposure Correction}

\author{Gehui Li}
\affiliation{%
  \institution{Dalian University of Technology}
  \city{}
  \country{}
}
\email{gehuili90@gmail.com}

\author{Jinyuan Liu}
\affiliation{%
  \institution{Dalian University of Technology}
  \city{}
  \country{}
}
\email{atlantis918@hotmail.com}

\author{Long Ma}
\affiliation{%
  \institution{Dalian University of Technology}
  \city{}
  \country{}
}
\email{malone94319@gmail.com}

\author{Zhiying Jiang}
\affiliation{%
  \institution{Dalian University of Technology}
  \city{}
  \country{}
}
\email{zyjiang0630@gmail.com}

\author{Xin Fan}
\affiliation{%
  \institution{Dalian University of Technology}
  \city{}
  \country{}
}
\email{xin.fan@dlut.edu.cn}

\author{Risheng Liu}
\authornote{Corresponding author: Risheng Liu.}
\affiliation{%
  \institution{Dalian University of Technology}
  \institution{Peng Cheng Laboratory}
  \city{}
  \country{}
}
\email{rsliu@dlut.edu.cn}

\renewcommand{\shortauthors}{Gehui Li et al.}

\begin{abstract}
Photographs taken with less-than-ideal exposure settings often display poor visual quality. Since the correction procedures vary significantly, it is difficult for a single neural network to handle all exposure problems. Moreover, the inherent limitations of convolutions, hinder the models ability to restore faithful color or details on extremely over-/under- exposed regions. To overcome these limitations, we propose a Macro-Micro-Hierarchical transformer, which consists of a macro attention to capture long-range dependencies, a micro attention to extract local features, and a hierarchical structure for coarse-to-fine correction. In specific, the complementary macro-micro attention designs enhance locality while allowing global interactions. The hierarchical structure enables the network to correct exposure errors of different scales layer by layer. Furthermore, we propose a contrast constraint and couple it seamlessly in the loss function, where the corrected image is pulled towards the positive sample and pushed away from the dynamically generated negative samples. Thus the remaining color distortion and loss of detail can be removed. We also extend our method as an image enhancer for low-light face recognition and low-light semantic segmentation. Experiments demonstrate that our approach obtains more attractive results than state-of-the-art methods quantitatively and qualitatively.
\end{abstract}



\begin{teaserfigure}
\centering
  \includegraphics[width=0.95\textwidth, height=0.15\textheight]{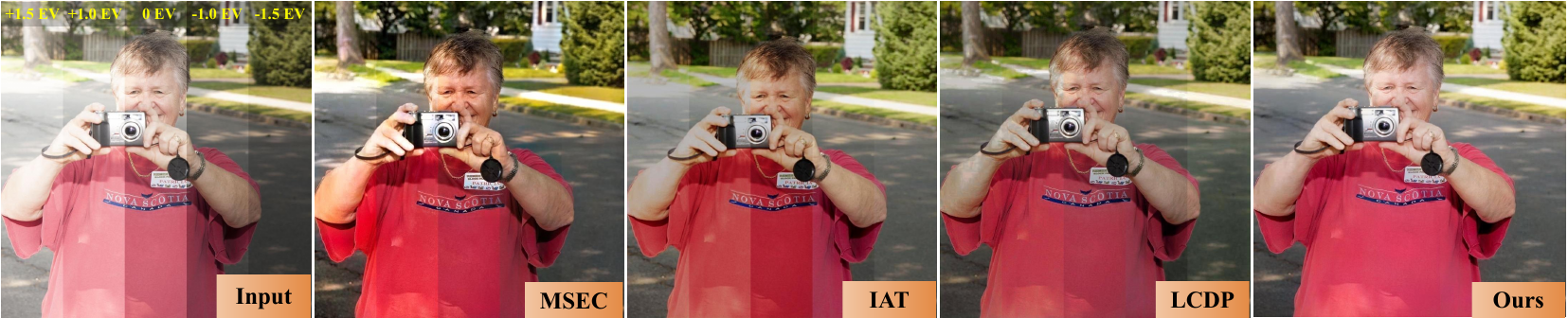}
  \caption{ The row displays a visual comparison of two cutting-edge methods alongside our proposed one, implemented on five separate exposures~(ranging from +1.5EV to -1.5EV). It is crucial to emphasize that our method not only precisely corrects  color distortions but also ensures consistency of the correction results.}
  \label{fig:teaser}
\end{teaserfigure}

\maketitle

\section{Introduction}
The ever-increasing interest in photography has intensified the need to capture expansive scenes under diverse lighting conditions. Nevertheless, image color and detail can be significantly compromised by sub-optimal and non-uniform illumination. While professional photographers often tackle exposure issues with specialized equipment and software, these solutions necessitate considerable expertise and can be costly. Moreover, harsh exposure conditions can adversely impact the performance of downstream computer vision tasks. Consequently, there is a pressing need for a model capable of correcting images with a wide range of exposure errors to achieve consistent, normal exposure without sacrificing color fidelity or detail.

Existing models \cite{guo2020zero, wang2019underexposed, wei2018deep, liu2022learning, liu2021retinex, ma2022low, ma2022practical, liu2022twin, jiang2022target, ma2023bilevel} primarily addressing single exposure error scenarios perform inadequately when confronted with images containing multiple exposure errors. Predicated on the assumption that scene lighting is low-light, these methods are inherently ill-equipped to adapt to varying exposure states. In addition, multi-exposure fusion methods\cite{liu2022target, liu2021learning, liu2020bilevel, liu2021smoa, liu2023holoco} are powerless when there is only one image of the same scene. These limitation have spurred the development of new approaches\cite{afifi2021learning, cui2022illumination, wang2022local, ma2022practical, eyiokur2022exposure} for tackling multiple exposure errors. 

MSEC\cite{afifi2021learning} addresses exposure correction of input images under diverse lighting. 
LCDP\cite{wang2022local} is designed to tackle the challenge of multiple exposure errors within a single image. 
However, existing approaches suffer from two critical shortcomings. First, when presented with a set of images with different exposure errors for the same scene, these methods fail to achieve a consistent correction state, leading to considerable discrepancies in the correction outcomes. Second, they struggle to restore distorted details in heavily overexposed and underexposed areas, resulting in unbalanced and lackluster colors in the corrected images.

Recognizing the importance of addressing the irreversible color distortion in severely overexposed or underexposed regions, we advocate for the utilization of richer semantic information from neighboring regions to compensate for the lost color. By providing the network with enhanced access to multi-scale information, the recovery of color-distorted regions is facilitated. To this end, we propose the Macro-Micro Transformer Restorer (MMT Restorer). Simultaneously, we introduce a dynamic negative sample generator and a contrast constraint function. By incorporating a substantial number of high-quality negative samples, the network learns a more consistent criterion, ultimately enhancing the consistency of input image correction. Furthermore, we present the Macro-Micro-Hierarchical Transformer (MMHT), a hierarchical extension of MMT Restorer, which refines the correction process from coarse to fine, further improving the quality of the corrected images.

The contributions of this paper are as follows:
\begin{itemize}
\item We propose a Macro-Micro-Hierarchical Transformer, which, in comparison to existing methods, yields visually appealing colors and consistently corrected outcomes.
\item Considering the limited capacity of convolutions to attend to non-local regions, we design a Macro-Micro Attention mechanism that effectively restores distorted color by strengthening global dependencies.
\item To explore the intrinsic information of over-/under- exposed image and the reference one, we raises a hierarchical contrastive learning scheme. In this manner, the restored image is pulled to the reference image and pushed away from the multiple exposed ones in a hierarchical latent feature spaces.
\item Extensive experimental results demonstrate our method empirically realizes the remarkable promotion in color and consistency of corrected images, as well as standard accuracy for low-light segmentation and detection compared with existing advanced networks.
\end{itemize}

\begin{figure*}[t]
\centering
\includegraphics[width=0.95\textwidth]{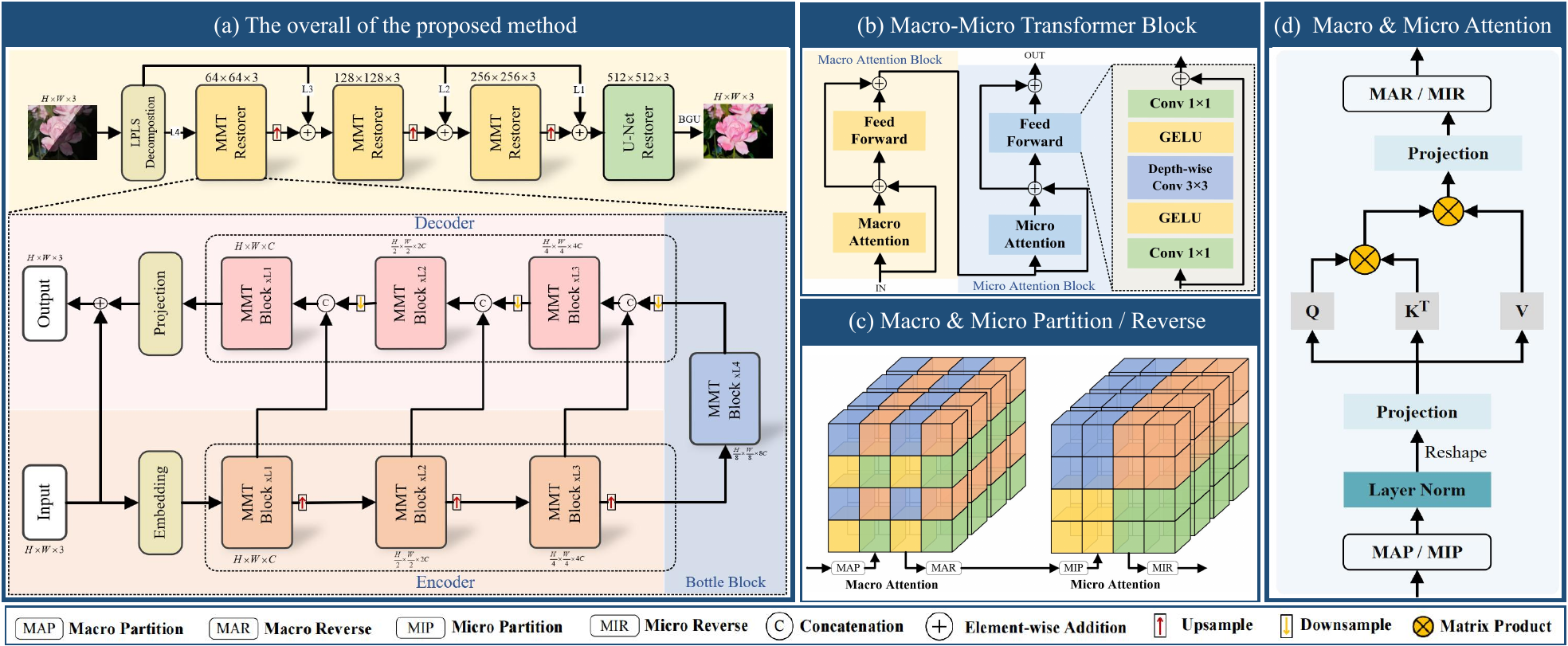} 
\caption{(a) The architecture of our MMHT method, which contains MMT Restorer and U-Net Restorer to  reconstruct the normal exposure image. (b) Macro-Micro Transformer Block with two complementary attention mechanisms to capture global dependencies. (c) Diagram of Micro \& Macro Attention Partition. (d) The structure of Micro \& Macro Attention.}
\label{msformer}
\end{figure*}

\section{Related Work}
\subsection{Methods against Harsh Lighting Condition}
\textbf{Low-Light Enhancement.}
Traditional methods\cite{guo2016lime, fu2016weighted, fu2016weighted, cai2017joint, zhang2018high, liu2012fixed, wu2019essential, zhang2019kindling, liu2021retinex, liu2020real} based on Retinex theory have been favored in recent years. These methods mainly decompose the image into reflectance map and illumination map, and then do processing on them separately to get the final result. Therefore, the goal is to predict the illumination map to obtain a well-exposed target image. Some deep learning-based methods\cite{eilertsen2017hdr, ignatov2017dslr, chen2018deep, wei2018deep, wang2019underexposed, FECNet, ECLNet, LowLightZhang, Huang_2023_CVPR, Huang_2022_CVPR, 9847530, li2022learning, liu2022attention} have also emerged subsequently. All these works are limited to image correction for underexposure conditions. However, mis-exposed images are common in reality, which also contain a large number of over-exposed and unevenly exposed images. Therefore, our goal is to construct a model capable of simultaneously correcting images with these different exposure levels.

\noindent\textbf{Exposure Correction.}
MSEC\cite{afifi2021learning} proposes a coarse-to-fine deep learning approach that focuses for the first time on the task of simultaneously correcting overexposed and underexposed images. IAT\cite{cui2022illumination} decomposes the Image Signal Processor (ISP) into local and global image components to correct the image from low-light or overexposure conditions and proceeds to solve the problem of low-light image semantic segmentation. LCDP\cite{wang2022local} proposes the LCDE module with Dual-Illumination mechanism to solve the problem of having multiple exposure errors on a single image. However, existing methods still have two common problems: inconsistent correction results and non-enjoyable colors. In this paper, we propose a Macro-Micro-Hierarchical Transformer to solve them.

\subsection{Vision Transformers}
A number of image restorers designed using transformer have appeared. For example, Uformer\cite{wang2022uformer} replaces the building block with its proposed LeWin Transformer Block based on the Encoder-Decoder structure of U-Net, which consists of window attention with a sliding window\cite{liu2021swin} and Locally Enhanced Feed-forward\cite{yuan2021incorporating}. Restormer\cite{zamir2022restormer} proposes Transformer Block consisting of A Multi-Dconv Head Transposed Attention with Gated Dconv Feed-forward Network. But these general image restorers, have a common problem that they do not adapt well to the exposure correction problem. In this paper, we propose a Macro-Micro Transformer Block containing two mechanisms of complementary Attention to solve this problem. Although Transformers has a stronger model fitting capability, it requires far more data than CNNs because of the lack of inductive bias. How to effectively combine the strengths of both is an important issue. A large number of networks\cite{wu2021cvt, yuan2021incorporating, zhang2022edgeformer, piao2019depth, piao2020a2dele, zhang2020select} designed for high level tasks in recent years have organically combined convolution and transformer. The Coatnet\cite{dai2021coatnet} is particularly enlightening in its exploration of the arrangement of convolution building blocks and transformer building blocks. Using a similar experimental method, we design the final network structure and its better adaptation to our task.

\section{Method}

\subsection{The overall of the proposed method}
Our proposed method commences with a Laplace decomposition to obtain four-layer decomposed images, L1-L4, derived from the input image. These decomposed images serve as the input terms for the skip-connection. The primary objective of this operation is to decompose the source images into distinct spatial frequency bands, thereby enabling the utilization of separate networks for the restoration of colors and details in specific frequency bands at varying decomposition levels. The L4 image is fed into the network as the initial term.
The architecture of the network comprises four stages, with stages 1, 2, and 3 designated for the Macro-Micro Transformer Restorer (MMT Restorer) and stage 4 for the U-Net Restorer. In the final stage, we employ a convolution-based structure as a substitute for the MMT Restorer. This decision is motivated by the fact that the transformer sacrifices its inductive bias while acquiring a global perceptual field through the attention structure. Consequently, we construct the U-Net Restorer by replacing the MMT Block with multi-layer convolutions, while preserving the overall structure of the MMT Restorer. In combination with convolution, our method achieves superior correction results. An additional benefit of this approach is that implementing convolution on larger sizes reduces the computation complexity.
Following each stage, up-sampling is performed to double the width and height. Ultimately, the $512\times512$ size image is up-sampled by BGU\cite{chen2016bilateral} to its original size, yielding the final output. A schematic representation of the framework is depicted in Figure \ref{msformer}.

\subsection{Macro-Micro Transformer Restorer}
As illustrated in Figure \ref{msformer} (a), the specific structure of our method is as follows. Given an input image, its dimensions are initially increased by an Embedding layer to C. Subsequently, the input is processed through an Encoder consisting of three stages, wherein identical MMT Blocks are stacked repeatedly with $L_{i}$ layers in each stage. Upon generating the feature output, the image size is downsampled using pooling, and the dimensions are increased via point-wise convolution. Following feature extraction by the Encoder, initial semantic features are obtained.
These semantic features are further refined using the Bottle Block, which serves as input for the Decoder to guide the generation of exposure-corrected images. Prior to feeding the image into the subsequent stage, it is upsampled using Pixel-shuffle\cite{shi2016real} and concatenated with the features processed by the Encoder at the same stage via a skip connection. In a similar way, the dimensionality is reduced to 3 by the Projection layer, and the resulting output is added to the input image to produce the exposure-corrected image.

\subsection{Macro-Micro Transformer Block}
In MMT Restorer, we alternately arrange Macro Attention and Micro Attention to obtain global and local features, but with linear complexity for spatial dimensions. Diagram is shown in Figure \ref{msformer} (b). The formulas of spatial attention is as follows. 
\begin{equation}
\begin{split}
\mathcal{A}_{Spatial}(\mathbf{Q}, \mathbf{K}, \mathbf{V}) &= Concat  \left(\right.  head  _{1}, \ldots , head_{\frac{C}{H_{d}}}\left)\right. ,\\
\end{split}
\end{equation}
$where\ \operatorname{head}_{i} = \operatorname{Softmax}\left(\frac{\mathbf{Q}_{i}\mathbf{K}_{i}^{T}}{\sqrt{H_{d}}}\right) \mathbf{V}_{i}$. $\mathbf{Q}_{i}=\mathbf{X}_{i} \mathbf{W}_{i}^{Q}, \mathbf{K}_{i}=\mathbf{X}_{i} \mathbf{W}_{i}^{K}$ and $\mathbf{V}_{i}=\mathbf{X}_{i} \mathbf{W}_{i}^{V} \in \mathbb{R}^{P \times H_{d}}$. $\mathbf{W}_{i}$ denotes the projection weights of the $i_{th}$ head for $\mathbf{Q}_{i}, \mathbf{K}_{i}, \mathbf{V}_{i}$. $P$ is is the number of total patches. $H_{d}$ is head dimension. $C$ is the total channel size.

\subsubsection{Micro Attention}
Conventional transformers employ global attention computation, resulting in a quadratic computational complexity. In contrast, Micro Attention confines attention computation to each window, effectively reducing the computational burden.
Micro Partition: $(\frac{H}{D} \times D, \frac{W}{D} \times D, C) \rightarrow (\frac{H W}{D^{2}}, D^{2}, C)$. After partition, we have the following formula.
\begin{equation}
\begin{split}
&\operatorname{MSA}_{Micro}(\mathbf{Q}, \mathbf{K}, \mathbf{V})=\left\{\mathcal{A}_{Spatial}(\mathbf{Q}_{j}, \mathbf{K}_{j}, \mathbf{V}_{j})\right\}_{j=0}^{\frac{HW}{D^2}},\\
\end{split}
\end{equation}
$where \ \mathbf{Q}_{j}, \mathbf{K}_{j}, \mathbf{V}_{j} \in \mathbb{R}^{D^2 \times C}$ are local window queries, keys, and values. Each window contains D $\times$ D patches.

Despite this significant reduction in computation, the exclusive focus on the window attention results in the loss of global information modeling capabilities. To address this limitation, we propose Macro Attention, characterized by a larger receptive field. This novel approach complements Micro Attention, augmenting the acquisition of multi-scale information.

\subsubsection{Macro Attention}
Drawing inspiration from the mechanism of dilated convolution, we incorporate it into the window division of attention in order to expand the receptive field. This adaptation effectively enhances the acquisition of global information while preserving linear complexity. Each window is inflated such that all windows fill the entire feature map without overlap.
Macro Partition: $(G \times \frac{H}{G}, G \times \frac{W}{G}, C) \rightarrow (G^{2}, \frac{H W}{G^{2}}, C )\rightarrow (\frac{H W}{G^{2}}, G^{2}, C )$. After partition, we have the following formula.
\begin{equation}
\begin{split}
&\operatorname{MSA}_{Macro}(\mathbf{Q}, \mathbf{K}, \mathbf{V})=\left\{\mathcal{A}_{Spatial}(\mathbf{Q}_{j}, \mathbf{K}_{j}, \mathbf{V}_{j})\right\}_{j=0}^{\frac{HW}{G^2}},\\
\end{split}
\end{equation}
$where \ \mathbf{Q}_{j}, \mathbf{K}_{j}, \mathbf{V}_{j} \in \mathbb{R}^{G^2 \times C}$ are dilated window queries, keys, and values. Each dilated window contains G $\times$ G patches.

\begin{table*}[t]
\centering
\caption{Quantitative comparisons on MSEC dataset. We compare each method with properly exposed reference image sets rendered by five expert photographers. The best results are highlighted with red. The second- and third-best results are highlighted in blue and green, respectively.}
\resizebox{0.91\textwidth}{!}{
\begin{tabular}{|l||>{\columncolor[HTML]{F0F0FF}}cc||>{\columncolor[HTML]{F0F0FF}}cc||>{\columncolor[HTML]{F0F0FF}}cc||>{\columncolor[HTML]{F0F0FF}}cc||>{\columncolor[HTML]{F0F0FF}}cc||>{\columncolor[HTML]{F0F0FF}}cc|}
\hline
\multicolumn{1}{|c||}{}                         & \multicolumn{2}{c||}{Expert A}                                                                        & \multicolumn{2}{c||}{Expert B}                                                                        & \multicolumn{2}{c||}{Expert C}                                                                        & \multicolumn{2}{c||}{Expert D}                                                                        & \multicolumn{2}{c||}{Expert E}                                                                        & \multicolumn{2}{c|}{Avg}   \\ 
\multicolumn{1}{|c||}{\multirow{-2}{*}{Method}} & \multicolumn{1}{c}{\cellcolor[HTML]{f0f0ff}PSNR} & \multicolumn{1}{c||}{SSIM} & \multicolumn{1}{c}{\cellcolor[HTML]{f0f0ff}PSNR} & \multicolumn{1}{c||}{SSIM} & \multicolumn{1}{c}{\cellcolor[HTML]{f0f0ff}PSNR} & \multicolumn{1}{c||}{SSIM} & \multicolumn{1}{c}{\cellcolor[HTML]{f0f0ff}PSNR} & \multicolumn{1}{c||}{SSIM} & \multicolumn{1}{c}{\cellcolor[HTML]{f0f0ff}PSNR} & \multicolumn{1}{c||}{SSIM} & \multicolumn{1}{c}{\cellcolor[HTML]{f0f0ff}PSNR} & \multicolumn{1}{c|}{SSIM} \\ \hline 
WVM\cite{fu2016weighted}                     & 14.488          & 0.788          & 15.803          & 0.699          & 15.117          & 0.678          & 15.863          & 0.693          & 16.469          & 0.704          & 15.548          & 0.688          \\
LIME\cite{guo2016lime}                    & 11.154          & 0.591          & 11.828          & 0.610          & 11.517          & 0.607          & 12.638          & 0.628          & 13.613          & 0.653          & 12.150          & 0.618          \\
HDR CNN w/PS\cite{eilertsen2017hdr}            & 15.812          & 0.667          & 16.970          & 0.699          & 16.428          & 0.681          & 17.301          & 0.687          & 18.650          & 0.702          & 17.032          & 0.687          \\
DPED (iPhone)\cite{ignatov2017dslr}           & 15.134          & 0.609          & 16.505          & 0.636          & 15.907          & 0.622          & 16.571          & 0.627          & 17.251          & 0.649          & 16.274          & 0.629          \\
DPED (BlackBerry)\cite{ignatov2017dslr}       & 16.910          & 0.642          & 18.649          & 0.713          & 17.606          & 0.653          & 18.070          & 0.679          & 18.217          & 0.668          & 17.890          & 0.671          \\
DPE (HDR)\cite{chen2018deep}               & 15.690          & 0.614          & 16.548          & 0.626          & 16.305          & 0.626          & 16.147          & 0.615          & 16.341          & 0.633          & 16.206          & 0.623          \\
DPE (S-FiveK)\cite{chen2018deep}           & 16.933          & 0.678          & 17.701          & 0.668          & 17.741          & 0.696          & 17.572          & 0.674          & 17.601          & 0.670          & 17.510          & 0.677          \\
RetinexNet\cite{wei2018deep}              & 10.759          & 0.585          & 11.613          & 0.596          & 11.135          & 0.605          & 11.987          & 0.615          & 12.671          & 0.636          & 11.633          & 0.607          \\
Deep UPE\cite{wang2019underexposed}                & 13.161          & 0.610          & 13.901          & 0.642          & 13.689          & 0.632          & 14.806          & 0.649          & 15.678          & 0.667          & 14.247          & 0.640          \\
Zero-DCE\cite{guo2020zero}                & 11.643          & 0.536          & 12.555          & 0.539          & 12.058          & 0.544          & 12.964          & 0.548          & 13.769          & 0.580          & 12.597          & 0.549          \\
RUAS\cite{liu2021retinex} & 10.166 & 0.391 & 10.522 & 0.440 & 9.356 & 0.411 & 11.013 & 0.441 & 11.574 & 0.466 & 10.526 & 0.430 \\
URetinex\cite{xu2020star}                & 11.420          & 0.632          & 12.230          & 0.700          & 11.818          & 0.672          & 13.078          & 0.701          & 14.066          & 0.735          & 12.522          & 0.688          \\
DALE\cite{kwon2020dale} & 13.294 & 0.691 & 14.324 & 0.757 & 13.734 & 0.722 & 14.256 & 0.743 & 14.511 & 0.763 & 14.024 & 0.735 \\
IAT(local)\cite{cui2022illumination}              & 16.610          & 0.750          & 17.520          & 0.822          & 16.950          & 0.780          & 17.020          & 0.780          & 16.430          & 0.789          & 16.910          & 0.783          \\
MSEC\cite{afifi2021learning}                    & 19.158          & 0.746          & 20.096          & 0.734          & 20.205          & 0.769          & 18.975          & 0.719          & 18.983          & 0.727          & 19.483          & 0.739          \\
IAT\cite{cui2022illumination}                     & \color[HTML]{009901}19.900          & \color[HTML]{0500FB}0.817          & \color[HTML]{009901}21.650          & \color[HTML]{0500FB}0.867          & \color[HTML]{009901}21.230          & \color[HTML]{009901}0.850          & \color[HTML]{009901}19.860          &\color[HTML]{0500FB} 0.844          & \color[HTML]{0500FB}19.340          & \color[HTML]{0500FB}0.840          & \color[HTML]{009901}20.340          & \color[HTML]{0500FB}0.844          \\
LCDPNet\cite{wang2022local}                 & \color[HTML]{0500FB}20.574          & \color[HTML]{009901}0.809          & \color[HTML]{0500FB}21.804          & \color[HTML]{009901}0.865          & \color[HTML]{0500FB}22.295          & \color[HTML]{0500FB}0.855          & \color[HTML]{0500FB}20.108          & \color[HTML]{009901}0.824          & \color[HTML]{009901}19.281          & \color[HTML]{009901}0.822          & \color[HTML]{0500FB}20.812          & \color[HTML]{009901}0.835          \\ 
\textbf{MMHT(Ours)}           & \color[HTML]{FE0000}\textbf{20.662} & \color[HTML]{FE0000}\textbf{0.816} & \color[HTML]{FE0000}\textbf{22.377} & \color[HTML]{FE0000}\textbf{0.869} & \color[HTML]{FE0000}\textbf{23.049} & \color[HTML]{FE0000}\textbf{0.865} & \color[HTML]{FE0000}\textbf{20.416} & \color[HTML]{FE0000}\textbf{0.831} & \color[HTML]{FE0000}\textbf{20.520} & \color[HTML]{FE0000}\textbf{0.833} & \color[HTML]{FE0000}\textbf{21.405} & \color[HTML]{FE0000}\textbf{0.843} \\ \hline
\end{tabular}}
\label{datatable}
\end{table*}

\subsection{Loss Function}
We train our model with following loss function:
\begin{equation}
\mathcal{L}_{\text {Total }}=\lambda_{1} \mathcal{L}_{\text {MAE}}+\lambda_{2} \mathcal{L}_{\text {DEC}}+\lambda_{3} \mathcal{L}_{\text {CR}},
\end{equation}
where $\mathcal{L}_{\text {MAE}}$ denotes Mean absolute loss, $\mathcal{L}_{\text {DEC}}$ the Pyramid decomposition loss and $\mathcal{L}_{\text {CR}}$ the contrastive regularization. 

\subsubsection{Mean absolute loss} $L_{1}$ loss is used to characterize the error between the corrected image and the ground truth.
\begin{equation}
\mathcal{L}_{\mathrm{MAE}}=\sum_{p=1}^{3H W}|\mathbf{Y}(p)-\mathbf{T}(p)|,
\end{equation}
where H, W are the height and width of the input size of the current image, respectively. $\mathbf{T}$ represents the ground truth and $\mathbf{Y}$ represents the network corrected image by bilateral interpolation to the original size image. p represents the pixels on the image. 

\subsubsection{Pyramid decomposition loss} This loss is used to measure the error between the decomposed corrected image and its ground truth layer by layer. These errors are proportionally merged together to form the loss of the decomposed image. 
\begin{equation}
\mathcal{L}_{\mathrm{DEC}}=\sum_{i=2}^{N} 2^{(i-2)} \sum_{p=1}^{3h_{i} w_{i}}\left|\mathbf{Y}_{(i)}(p)-\mathbf{T}_{(i)}(p)\right|,
\end{equation}
where N represents the number of levels of the pyramid. $\mathbf{Y}_{(i)}$ stands for i-th corrected pyramid output. $\mathbf{T}_{(i)}$ represents the i-th gaussian pyramid layer decomposed from $\mathbf{T}$. $h_{i}=\frac{1}{2^{i-1}} h $ and $ w_{i}=\frac{1}{2^{i-1}} w$.

\begin{figure}[h]
\centering
\includegraphics[width=0.95\columnwidth]{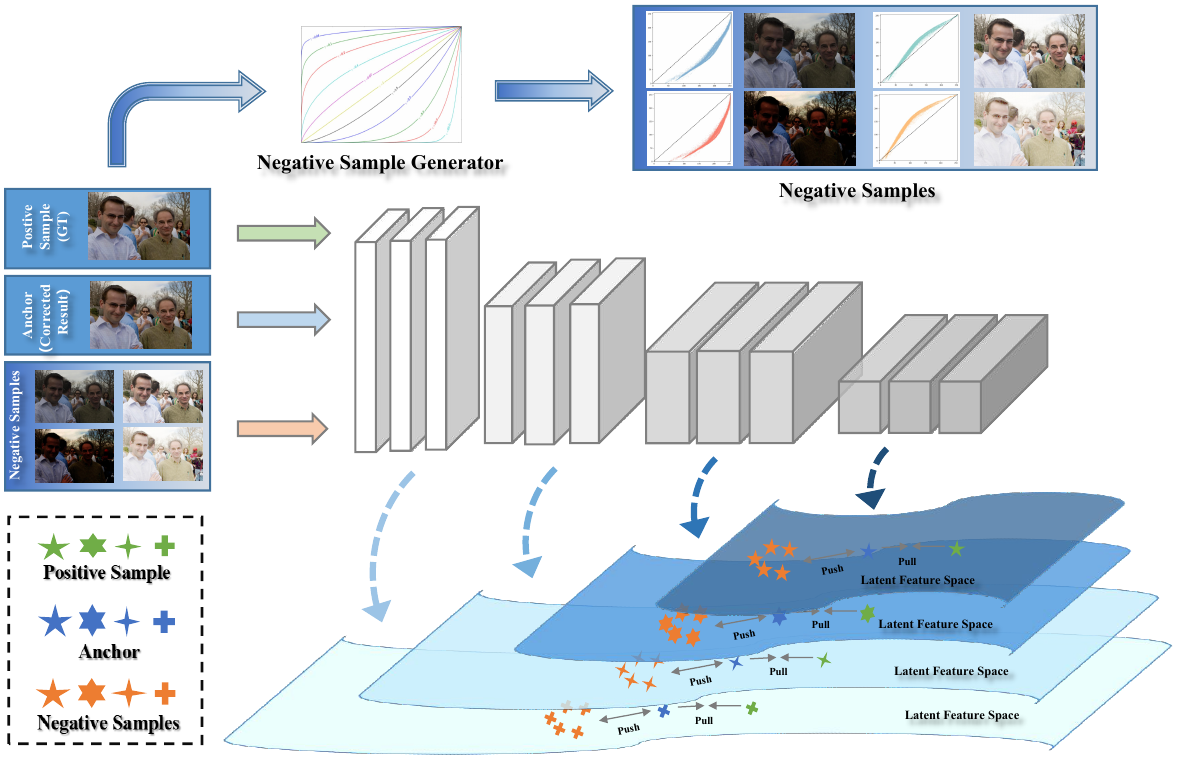} 
\caption{Adaptation Contrastive Regularization with Dynamic Negative Sample Generator}
\label{cr}
\end{figure}

\begin{figure*}[t]
\centering
\includegraphics[width=0.91\textwidth]{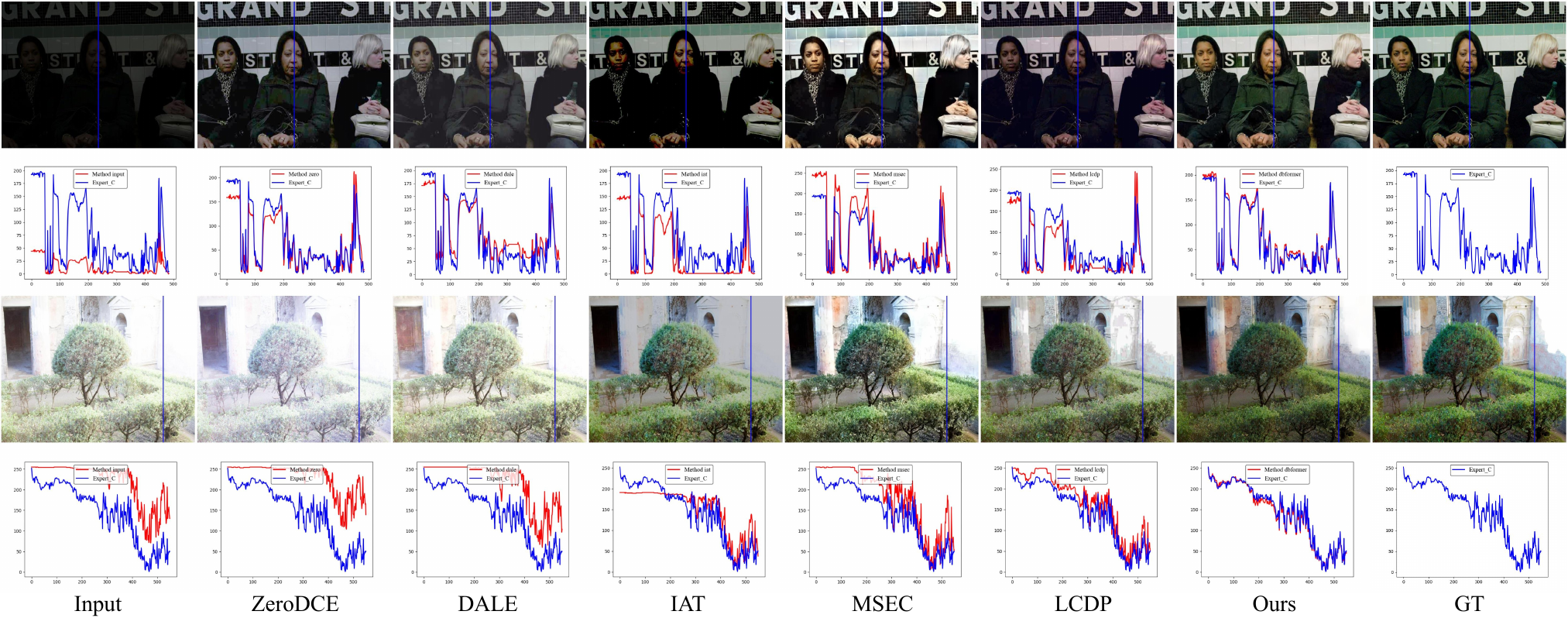} 
\caption{Qualitative comparisons on MSEC dataset. The blue line shows that the pixel values of the column are converted to perceived brightness and compared with that of ground truth.}
\label{finalshow}
\end{figure*}

\subsubsection{Contrastive regularization} 
As highlighted in existing contrast learning methods\cite{chen2020simple}, diverse negative samples are vital for effective contrast learning. We propose a dynamic negative sample generation strategy, wherein for each input GT image, K negative samples are generated via gamma correction, with gamma values uniformly selected from the range [G1, G2]. Training with multiple negative samples enhances the model's generalizability to unseen exposure error images, crucial for improving performance and real-world applicability.

As shown in the Figure \ref{cr}, we formulate a contrastive loss to minimize the distance between the correction result and the positive sample while maximizing the distance from the negative sample. A pre-trained network, R, is employed to extract features and apply contrastive regularization. The rich information obtained through feature extraction enables the contrastive loss to effectively contribute to the task.

\begin{equation}
\mathcal{L}_{\mathrm{CR}}=\frac{\sum_{s=1}^{S} D\left(f_{s}, f_{s}^{+}\right)}{\sum_{k=1}^{K}\sum_{s=1}^{S} D\left(f_{s}, f_{ks}^{-}\right)},
\end{equation}
where $D(x, y)=\|x-y\|_{1}$, $f_{s}=R(Y)$, $f_{s}^{+}=R(T)$, $f_{s}^{-}=R(F)$, $s=1,2, \ldots, S$. K represents the number of samples generated by the negative sample generator. Both positive and negative samples are passed through the pre-trained VGG19 network, and the corrected image is downsampled S times. Y represents the corrected image, T represents the ground truth and F represents the negative images generated by Dynamic Negative Sample Generator.

\section{Experiment}

\begin{figure*}[t]
\centering
\includegraphics[width=0.95\textwidth]{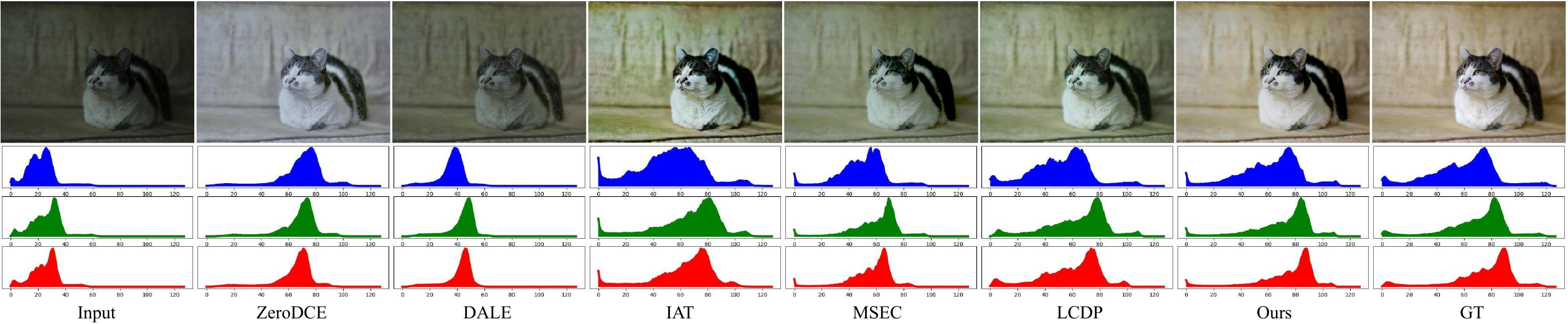} 
\caption{Compare the color performance of our method with other SOTA methods. Below the image is the RGB histogram.}
\label{color}
\end{figure*}
\begin{figure*}[t]
\centering
\includegraphics[width=0.95\textwidth]{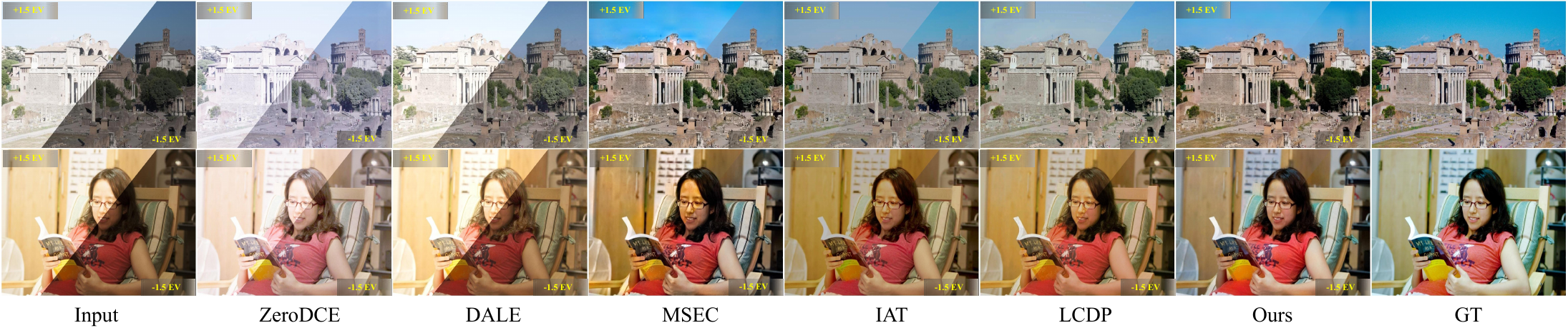} 
\caption{The consistency of correction results between our method and existing methods is compared. The correction results of overexposure and underexposure are concatenated.}
\label{corshow}
\end{figure*}

\subsection{Implementation Details}
We use the ADAM to optimize the network for 150 epochs with a initial learning rate of 1e-4. 
The weight of the loss function is set to $\lambda_{1} = 1.0 $, $\lambda_{2} = 1.0 $ and $\lambda_{3} = 0.6$. 
The level of the Laplacian pyramid decomposition is $N=4$. The lowest gamma value is $G1 = 0.3$ and the highest gamma value is $G2 = 2.8$. 
The number of negative samples generated is $K = 6$. The number of feature maps obtained by the feature extractor is $S = 4$. 
Before training, we do data augmentation and resize the image to $512 \times 512$ $(h \times w)$. We conduct all experiments with an NVIDIA 3090TI GPU.

\begin{table}[h]
\centering
\caption{Quantitative comparisons of color performance on MSEC and LCDP datasets.}
\resizebox{0.8\columnwidth}{!}{
\begin{tabular}{|l|cccccc|}
\hline
Methods(MSEC) & PNSR$\uparrow$ & SSIM$\uparrow$  & FID$\downarrow$ & CF$\uparrow$    & $\Delta$ CF$\downarrow$    & NIQE$\downarrow$ \\
\hline
RUAS & 9.356 & 0.411 & 30.125 & 22.612 & 19.792 & 4.643 \\
URetinex & 11.818 & 0.672 & 27.451 & 28.597 & 13.807 & \color[HTML]{0500FB}2.654 \\
MSEC & 20.205 & 0.769 & 20.264 & \color[HTML]{0500FB}38.898 & \color[HTML]{0500FB}3.506 &\color[HTML]{009901} 2.838 \\
IAT & \color[HTML]{009901}21.230 & \color[HTML]{009901}0.850 & \color[HTML]{0500FB}12.091 & \color[HTML]{009901}37.188 & \color[HTML]{009901}5.216 & 3.132 \\
LCDP & \color[HTML]{0500FB}22.295 & \color[HTML]{0500FB}0.855 & \color[HTML]{009901}12.752 & 33.293 & 9.111 & 2.914 \\
(MMHT)Ours & \color[HTML]{FE0000}\textbf{23.049} & \color[HTML]{FE0000}\textbf{0.865} & \color[HTML]{FE0000}\textbf{10.223} & \color[HTML]{FE0000}\textbf{40.202} & \color[HTML]{FE0000}\textbf{2.202} & \color[HTML]{FE0000}\textbf{2.339} \\
\hline \hline
Methods(LCDP) & PNSR$\uparrow$ & SSIM$\uparrow$  & FID$\downarrow$ & CF$\uparrow$    & $\Delta$ CF$\downarrow$    & NIQE$\downarrow$ \\
\hline
LIME & 17.335 & 0.686 & 47.725 & 32.949 & 6.077 & 4.248 \\
IAT & 18.842 & 0.684 & \color[HTML]{009901}37.516 & 32.143 & 6.883 & \color[HTML]{009901}2.975 \\
RetinexNet & 19.250 & 0.704 & 52.723 & \color[HTML]{0500FB}33.541 &\color[HTML]{0500FB} 5.485 & 4.947 \\
MSEC & \color[HTML]{009901}20.377 & \color[HTML]{009901}0.779 & 45.231 & 29.005 & 10.021 & 3.841 \\
LCDP & \color[HTML]{0500FB}23.239 & \color[HTML]{0500FB}0.842 & \color[HTML]{0500FB}28.914 & \color[HTML]{009901}32.975 & \color[HTML]{009901}6.051 & \color[HTML]{0500FB}2.354 \\
(MMHT)Ours & \color[HTML]{FE0000}\textbf{23.715} & \color[HTML]{FE0000}\textbf{0.855} & \color[HTML]{FE0000}\textbf{25.147} & \color[HTML]{FE0000}\textbf{34.753} & \color[HTML]{FE0000}\textbf{4.273} & \color[HTML]{FE0000}\textbf{2.142} \\
\hline
\end{tabular}}
\label{mseccol}
\end{table}

\subsection{Comparisons with State-of-the-Art Methods}
We evaluate our proposed method by comparing it with existing SOTA methods. First, we examine quantitative metrics, employing the widely used peak signal-to-noise ratio (PSNR) and structural similarity (SSIM) as our evaluation indicators. To assess color performance, we utilize the Fréchet Inception Distance (FID)\cite{heusel2017gans} to measure the similarity between two sets of images, the Colorfulness Score (CF)\cite{hasler2003measuring} to gauge the richness of image colors, and additional metrics such as $\Delta$CF and NIQE. Second, we conduct visual comparisons of exposure correction results to further substantiate the effectiveness of our method.

\subsubsection{Quantitative Comparisons}
\begin{figure}[h]
\centering
\includegraphics[width=0.85\columnwidth]{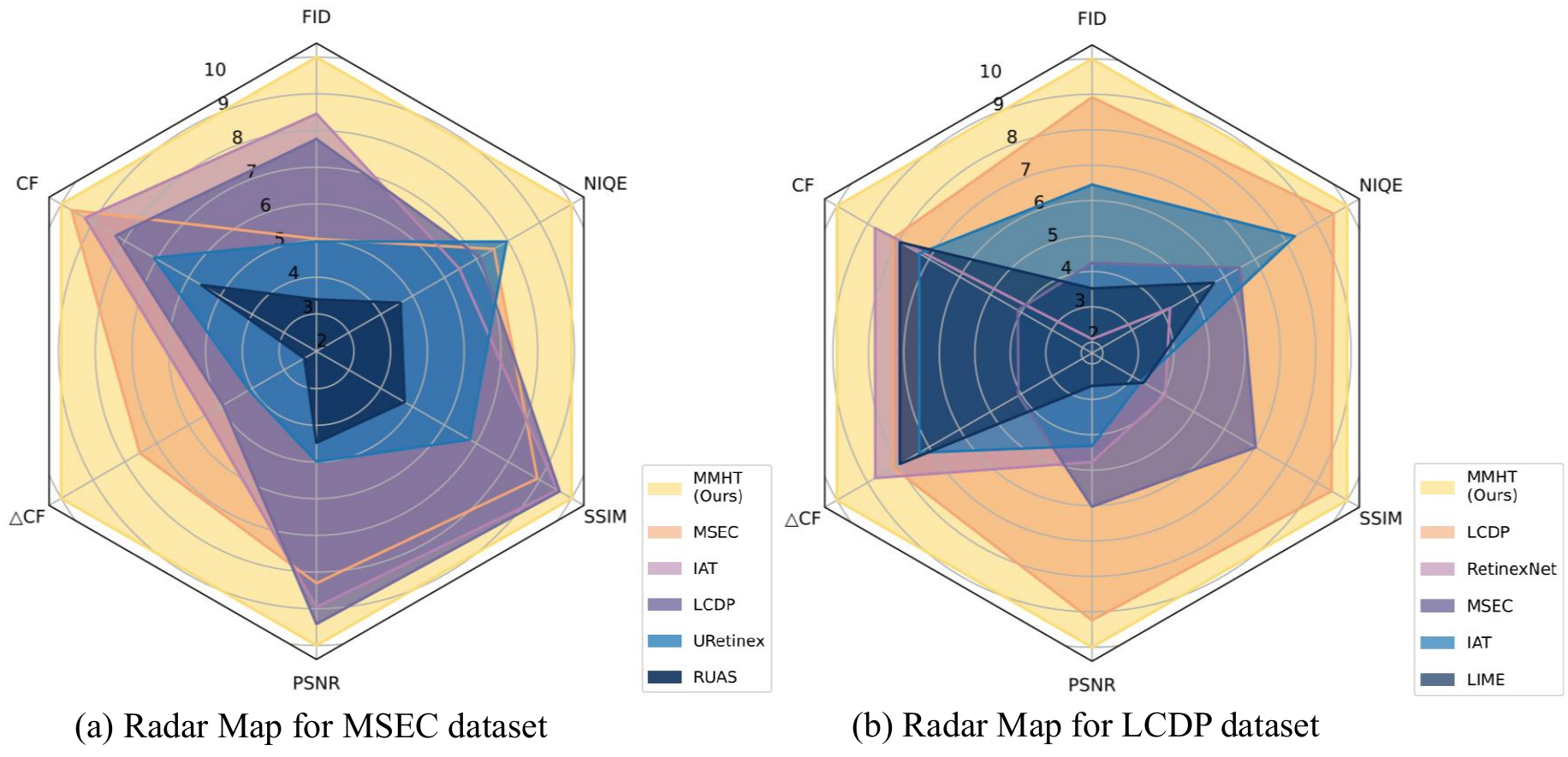}
\caption{Radar Map for Performance Comparisons.}
\label{radar}
\vspace{-9pt}
\end{figure}

The test set of MSEC contains 1181 sets of images of the same scene with five different exposures, accompanied by the correction results of five different experts.
Table \ref{datatable} shows that our method outperforms all existing methods in terms of PSNR and SSIM on all five expert test sets of MSEC. For color performance comparisons in Table \ref{mseccol}, our method achieves the lowest FID, indicating that the proposed method is capable of producing realistic, natural and consistent images. In addition, the lowest NIQE indicates that the corrected images of our method have the best subjective viewability. There are some methods that are closer to our method on CF, but the FIDs of these methods are much higher than ours, which means that they generate a lot of rare colors. Therefore, $\Delta$CF is used to measure the difference between the CF of the corrected image and that of the expert image. A lower $\Delta$CF indicates a more realistic color of the image. Our $\Delta$CF is significantly better than all other methods. 
The performance comparisons are presented in the Figure \ref{radar} in the form of radar map. 
Each image of the LCDP dataset has both overexposed and underexposed areas. Our method also gains SOTA on LCDP dataset, which shows that the images corrected by our method are in excellent consistency.

\begin{figure}[h]
\centering
\includegraphics[width=0.85\columnwidth]{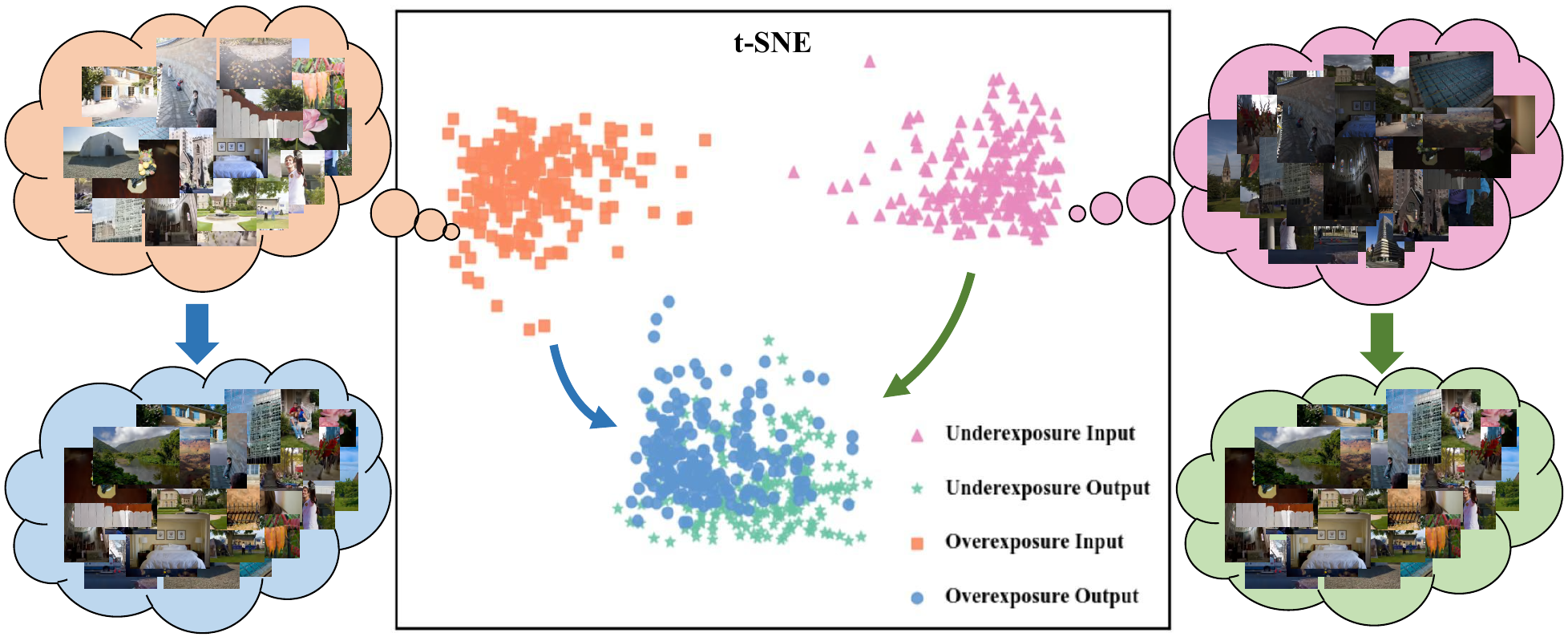}
\caption{t-SNE visualization to show correction consistency.}
\label{tsne}
\vspace{-9pt}
\end{figure}

\subsubsection{Qualitative Comparisons}
Figure \ref{finalshow} presents a comparison of image quality between our method and SOTA methods. For both underexposed and overexposed images, our method accurately corrects them to the proper exposure state. In Figure \ref{corshow}, we evaluate the consistency of correction results by comparing corrected images produced by our method with those corrected by existing methods. Our method's results exhibit remarkable consistency, displaying not only similar color and brightness but also eliminating differences in outcomes due to input variations. Furthermore, as illustrated in Figure \ref{color}, images corrected by our method boast both the most visually appealing color performance and superior detail preservation. In contrast, other methods either introduce inconsistent lighting or generate color distortion. Furthermore, as can be seen in Figure \ref{tsne}, after being processed by MMHT, the underexposure and overexposure representations tend to be intersected together.

\begin{figure*}[t]
\centering
\includegraphics[width=0.95\textwidth]{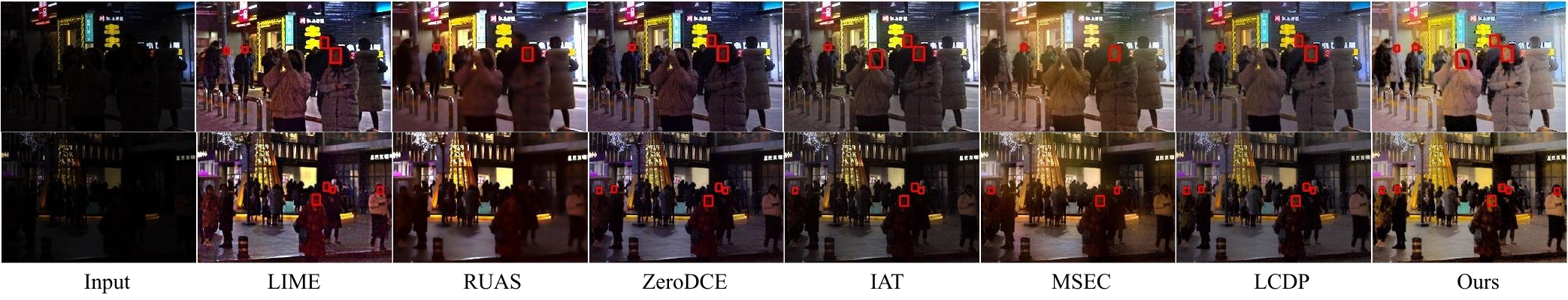}
\caption{Detection results on DarkFace dataset.}
\label{hightwo1}
\end{figure*}

\begin{figure*}[t]
\centering
\includegraphics[width=0.95\textwidth]{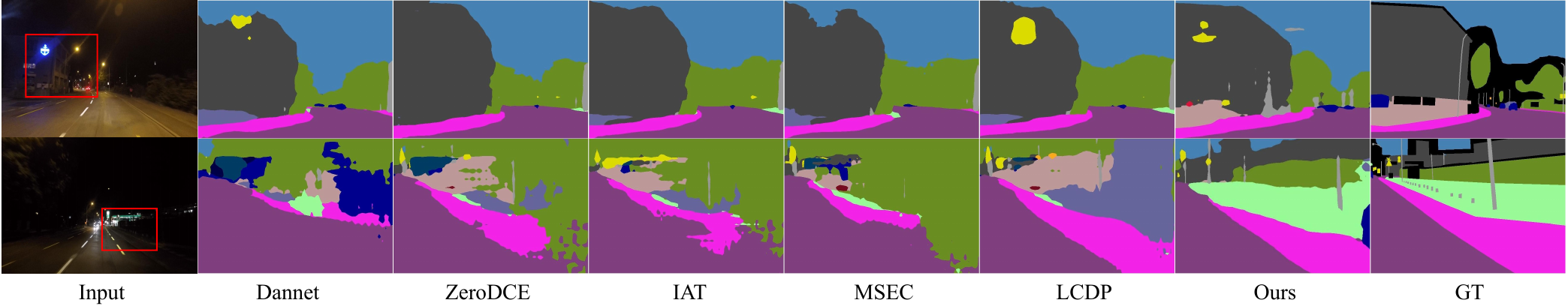}
\caption{Segmentation results on ACDC-Nighttime dataset.}
\label{hightwo2}
\end{figure*}

\begin{figure}[h]
\centering
\includegraphics[width=0.95\columnwidth]{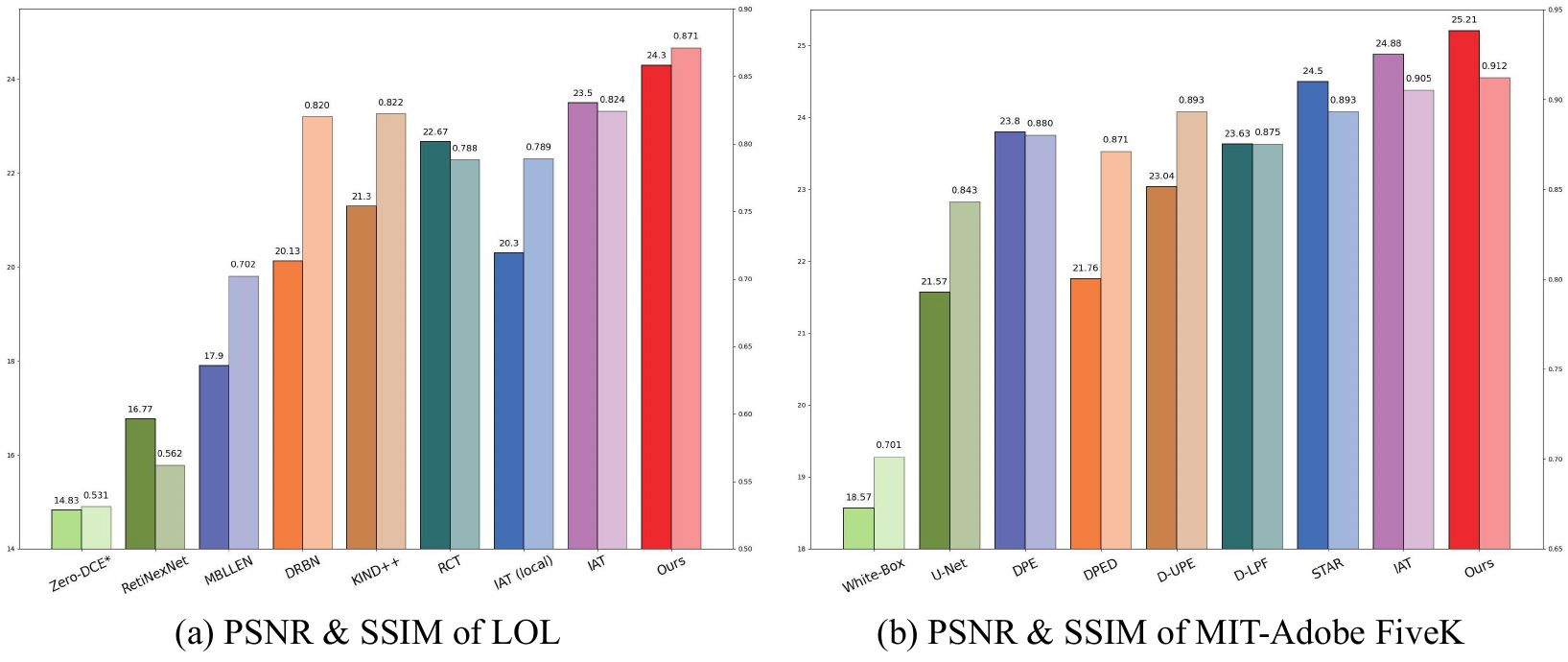}
\caption{Quantitative comparisons on LOL and MIT-5K.}
\label{lowtwo}
\end{figure}

\begin{table*}[t]
\small
\centering
\caption{Performance comparisons on high-level tasks. We retrain the detector/segmentator in all cases containing the enhancer.}
\resizebox{0.95\textwidth}{!}{
\begin{tabular}{|c|c|c|c|c|c|c|c|c|c|c|}
\hline
Task & \multicolumn{3}{c|}{Dark Face Detector} & \multicolumn{7}{c|}{Enhancer + Detector (Finetune)}  \\
\hline
Method & HLA\cite{wang2022unsupervised}          & REG\cite{liang2021recurrent}         & MEAT\cite{cui2021multitask}    &LIME\cite{guo2016lime}   & ZeroDCE\cite{guo2020zero} & MSEC\cite{afifi2021learning}  & RUAS\cite{liu2021retinex}  & LCDP\cite{wang2022local}  & IAT\cite{cui2022illumination}   & MMHT(Ours) \\
\hline
mAP & 0.607       & 0.514       & 0.526    &0.644  & 0.665   & 0.659 & 0.642 & 0.654 & 0.663 & 0.675     \\
\hline \hline
Task & \multicolumn{3}{c|}{Nighttime Semantic Segmentator} & \multicolumn{7}{c|}{Enhancer + Segmentator
 (Finetune)} \\
\hline
Method & DANNet\cite{xu2020star}         & CIC\cite{lengyel2021zero}                              & GPS-GLASS\cite{lee2022gps}      &LIME\cite{guo2016lime}                & ZeroDCE\cite{guo2020zero}  & MSEC\cite{afifi2021learning}   & RUAS\cite{liu2021retinex}  & LCDP\cite{wang2022local}  & IAT\cite{cui2022illumination}   & MMHT(Ours) \\ \hline
mIoU & 0.398          & 0.264         & 0.380        &0.447     & 0.452    & 0.449  & 0.448 & 0.455 & 0.456 & 0.461 \\ \hline   
\end{tabular}}
\label{highdata}
\end{table*}

\subsection{Image Enhancement Results}
For LOL dataset\cite{wei2018deep}, the compared methods include
Zero-DCE\cite{guo2020zero}, RetiNexNet\cite{wei2018deep}, MBLLEN\cite{lv2018mbllen}, DRBN\cite{yang2020fidelity}, KIND++\cite{zhang2021beyond}, RCT\cite{kim2021representative}, IAT\cite{cui2022illumination}.
For MIT-Adobe FiveK dataset\cite{bychkovsky2011learning}, the compared methods include
White-Box\cite{hu2018exposure}, U-Net\cite{ronneberger2015u}, DPE\cite{chen2018deep}, DPED\cite{ignatov2017dslr}, D-UPE\cite{wang2019underexposed}, D-LPF\cite{moran2020deeplpf}, STAR\cite{zhang2021star}, IAT\cite{cui2022illumination}. The quantitative comparisons with existing methods are shown in Figure \ref{lowtwo}, which proves that our method also performs superior in low light scenes.

\subsection{High-level Vision Applications}
\begin{figure*}[t]
\centering
\includegraphics[width=0.98\textwidth]{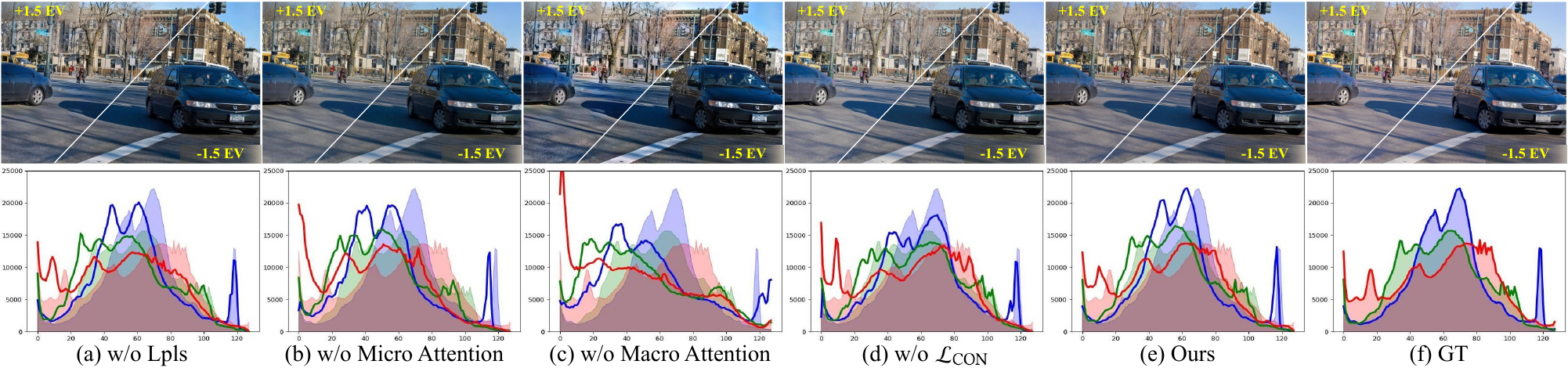}
\caption{Visualization results of ablation experiments are shown. The GT images are filled RGB curves while the ablation experimental images are unfilled RGB curves.}
\label{abexpimg}
\end{figure*}

To demonstrate the generalizability of our method, we apply it to relevant high level vision tasks, including low-light face and low-light semantic segmentation tasks.
 
Here we adopt the DARKFACE\cite{yang2020advancing} and ACDC\cite{sakaridis2021acdc} datasets for evaluating low-light detection and segmentation respectively.
To fully evaluate its performance, we not only compare some of the correction methods, but also consider specific detection methods and segmentation methods. The quantitative comparisons with existing methods are shown in Figure \ref{hightwo1} and \ref{hightwo2} while quantitative results are in the Table \ref{highdata}. Both quantitatively and quantitatively demonstrate the significance of our method for downstream tasks.

\subsection{Ablation Study}

\subsubsection{Experiment on Transformer and Convolution Arrangement and the Effect of Laplacian Pyramid Decomposition}
To explore the optimal arrangement of transformer and convolution, we design an ablation study for the network used in each stage, with results in Table \ref{abexpdt}. As U-Net is progressively replaced by the MMT Restorer, the quantitative results improve. However, there is a slight drop in PSNR and SSIM when the last network is replaced, as removing the convolution entirely eliminates the inductive bias. Combining convolution with transformers yields the better results, with R1, R2, R3 as the best pipeline. We also conduct ablation experiments on Laplacian pyramid decomposition. As shown in the Table \ref{abexpdt} and Figure \ref{abexpimg}(a), MMT Restorer performs well even without Laplacian pyramid decomposition but compromises the coarse-to-fine correction, negatively impacting results qualitatively and quantitatively.

\begin{table}[h]
\centering
\caption{Exploring Transformer and Convolution Arrangement and the effect of Laplacian Pyramid Decomposition. R1 means replacing the U-Net Restorer with MMT Restorer in stage 1 of the pipeline.}
\resizebox{0.8\columnwidth}{!}{
\begin{tabular}{cccccccccc}
 \toprule
R1         & R2         & R3         & R4         & Lpls       & PSNR$\uparrow$    & SSIM $\uparrow$ & $\Delta$ CF$\downarrow$     & FID$\downarrow$     \\ \midrule
-         & -         & -         & -         & \ding{55} & 22.785 & 0.853 & 2.119 & 10.331 \\ \midrule
\ding{55} & \ding{55} & \ding{55} & \ding{55} & \ding{51} & 20.517 & 0.771 & 3.419 & 20.964 \\
\ding{51} & \ding{55} & \ding{55} & \ding{55} & \ding{51} & 21.875 & 0.811 & 3.331 & 15.397 \\
\ding{51} & \ding{51} & \ding{55} & \ding{55} & \ding{51} & 22.269 & 0.831 & 2.798 & 13.778 \\
\ding{51} & \ding{51} & \ding{51} & \ding{55} & \ding{51} & \textbf{23.049} & \textbf{0.865} & \textbf{2.202} & \textbf{10.223} \\
\ding{51} & \ding{51} & \ding{51} & \ding{51} & \ding{51} & 22.971 & 0.859 & 2.229 & 10.414 \\ \bottomrule
\end{tabular}}
\label{abexpdt}
\end{table}

\subsubsection{Experiments on the Effect of Macro-Micro Attention}
Global information acquisition is crucial for addressing image exposure correction issues. 
However, the micro attention restricts operations to a single window and loses long-range dependencies. 
We introduce macro attention with a larger receptive field in combination with micro attention to effectively acquire multi-scale information. 
An ablation study is designed for macro-micro attention, with results presented in Table \ref{abexpdt2}.
Replacing either micro or macro attention with the other results in a significant increase in FID and CF with a decrease in PSNR and SSIM. Furthermore, swapping their order also inflicts a slight detriment to these metrics. In the Figure \ref{abexpimg}(c), we observe that omitting macro attention severely impacts the results, causing substantial color distortions and artifacts.

\subsubsection{Experiment on the Effect of Contrastive Regularization}
To examine the effect of contrastive loss, we remove the contrastive loss in the adjusting attention order. As shown in the Table \ref{abexpdt2}, there is a significant reduction in the PSNR and SSIM. 
Figure \ref{abexpimg}(d) shows that removing contrastive learning leads to inconsistent correction results for different exposure inputs.
Contrastive constraint encourages the model to learn features distinguishing well-exposed images from exposure-error images. Using abundant negative samples forces the model to learn more discriminative features, resulting in consistent exposure correction performance. 

\begin{table}[h]
\centering
\caption{Exploring the effect of Macro-Micro Attention and Contrastive Regularization.}
\resizebox{0.85\columnwidth}{!}{
\begin{tabular}{ccccccccc}
\toprule
Mic          & Mac          & Mic-Mac        & Mac-Mic        & $L_{CR}$   & PSNR$\uparrow$            & SSIM$\uparrow$           & $\Delta$ CF$\downarrow$     & FID$\downarrow$     \\ \midrule
\ding{51}          & \ding{55}          & \ding{55}          & \ding{55}          & \ding{55}          & 22.092          & 0.823          & 3.945          & 16.712          \\
\ding{55}          & \ding{51}          & \ding{55}          & \ding{55}          & \ding{55}          & 22.325          & 0.836          & 2.616          & 12.301          \\
\ding{55}          & \ding{55}          & \ding{51}          & \ding{55}          & \ding{55}          & 22.791          & 0.843          & 2.419          & 11.818          \\
\ding{55}          & \ding{55}          & \ding{55}          & \ding{51}          & \ding{55}          & 22.901          & 0.851          & 2.305          & 10.597          \\ \midrule
\ding{51}          & \ding{55}          & \ding{55}          & \ding{55}          & \ding{51}          & 22.225          & 0.835          & 3.978          & 16.645          \\
\ding{55}          & \ding{51}          & \ding{55}          & \ding{55}          & \ding{51}          & 22.494          & 0.846          & 2.511          & 12.219          \\
\ding{55}          & \ding{55}          & \ding{51}          & \ding{55}          & \ding{51}          & 22.949          & 0.862          & 2.415          & 11.774          \\
\textbf{\ding{55}} & \textbf{\ding{55}} & \textbf{\ding{55}} & \textbf{\ding{51}} & \textbf{\ding{51}} & \textbf{23.049} & \textbf{0.865} & \textbf{2.202} & \textbf{10.223} \\ \bottomrule
\end{tabular}}
\label{abexpdt2}
\end{table}

\begin{figure}[h]
\centering
\includegraphics[width=0.98\columnwidth]{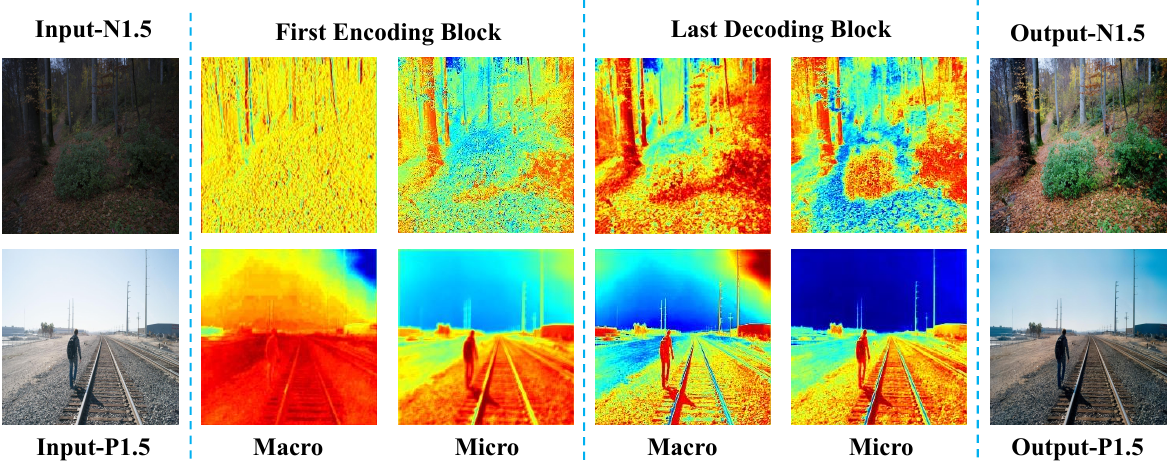} 
\caption{Visualization of heat maps from the first encoding block and last decoding block.}
\label{heatmap}
\end{figure}

\subsection{Feature map visualization}
To further explain the role of multiple attention and hierarchical network structure, we visualize the feature map after processing of the first block of the encoder in stage 1 and the last block of the decoder in stage 3. The results are presented in the Figure \ref{heatmap}. 
Long-range dependencies appear after macro attention, while short-range dependencies appear after micro attention. This aids the network in adapting different adjustments based on varying semantics.

\section{Conclusion}
In this paper, we address the issues of inconsistent results and color distortion in exposure correction tasks. 
To model long-range dependencies, a Macro-Micro attention mechanism has been developed, which significantly enhances image color and detail. 
A contrastive constraint is then established to effectively resolve inconsistent correction results. By integrating the advantages of both transformer and convolution, a four-stage pipeline is constructed, which further improves the quality of corrected images.
Our proposed model is assessed on multiple challenging exposure-error datasets and applied as an enhancer for high-level tasks. Results demonstrate that our approach significantly outperforms state-of-the-art methods, indicating its potential applicability in exposure-related tasks.

\begin{acks}
This work is partially supported by the National Key R\&D Program of China (No. 2022YFA1004101), the National Natural Science Foundation of China (No. U22B2052).
\end{acks}

\balance
\bibliographystyle{ACM-Reference-Format}
\bibliography{sample-base}

\appendix

\end{document}